\pdfoutput=1
\documentclass[11pt]{article}

\usepackage{acl}

\usepackage{times}
\usepackage{latexsym}

\usepackage[T1]{fontenc}
\usepackage[utf8]{inputenc}

\usepackage{microtype}
\usepackage{url,color}
\usepackage{amsmath}
\usepackage{verbatim}
\usepackage{subcaption}
\usepackage{booktabs}
\usepackage{bm}
\usepackage{graphicx}
\usepackage{natbib}
\usepackage{multirow,array}
\usepackage{amssymb}
\usepackage{tikz}

\usepackage{enumerate}

\newcommand{\secref}[1]{Section \ref{#1}}
\newcommand{\figref}[1]{Figure \ref{#1}}

\newcommand{\tabref}[1]{Table \ref{#1}}

\newcommand{\cut}[1]{}

\title{Few-Shot Natural Language Inference Generation with PDD:\\Prompt and Dynamic Demonstration}

\author{Kaijian Li, Shansan Gong, Kenny Q. Zhu \\
School of Electronic Information and Electrical Engineering, Shanghai Jiao Tong University \\ {2275135452@qq.com, gongshansan@sjtu.edu.cn, kzhu@cs.sjtu.edu.cn}}

\begin{document}
\maketitle
\begin{abstract}
Natural Language Inference Generation task is to generate a text hypothesis 
given a text premise and a logical relation between the two. This task can be used in data augmentation and controllable text generation in practice. 
In this paper, we propose language models with prompt and dynamic demonstration (LM-PDD) to tackle this problem in few-shot settings. 
Our framework outperforms standard fine-tuned models with low 
resource, achieving an average 8\% absolute improvement on SNLI and MNLI 
datasets, and the results on 13 natural language classification tasks also
show that our dynamic demonstration method has good generalizability.      
\end{abstract}

\section{Introduction}
Natural Language Inference (NLI), also known as the Recognizing Textual 
Entailment (RTE) is a task of determining whether a text premise $p$ 
can entail, contradict or be neutral with a given hypothesis $h$~\citep{DBLP:conf/mlcw/DaganGM05}. 
This task servers as an important benchmark for testing model's reasoning 
ability.
% It's also widely used as an invaluable aid in other NLP tasks, such as dialogue consistency~\citep{DBLP:conf/acl/WelleckWSC19}.

Although NLI is proposed as a classification problem, 
it's also worth considering it as a generation task. Recently, 
several works reformulate NLI task as a generation task~\citep{DBLP:journals/corr/KolesnykRR16,DBLP:journals/csl/StarcM17,DBLP:journals/corr/abs-1803-02710,DBLP:journals/corr/abs-1909-09788}. 
They explore how to generate the hypothesis $h$ given 
a premise $p$ and a logic condition $c$ or reversely generate $p$ given $h$ and $c$. 
\citet{DBLP:journals/corr/KolesnykRR16} show that the task of generating 
entailed sentences can be served as a benchmark to measure the reasoning 
ability of sequence to sequence models.
\citet{DBLP:journals/csl/StarcM17} first use it for data augmentation. They use generated hypotheses to construct a new NLI style dataset, on which a better classifier can 
be trained.

\begin{figure}[t]
	\centering
	\scalebox{1.0}{\includegraphics[width=1.0\columnwidth]{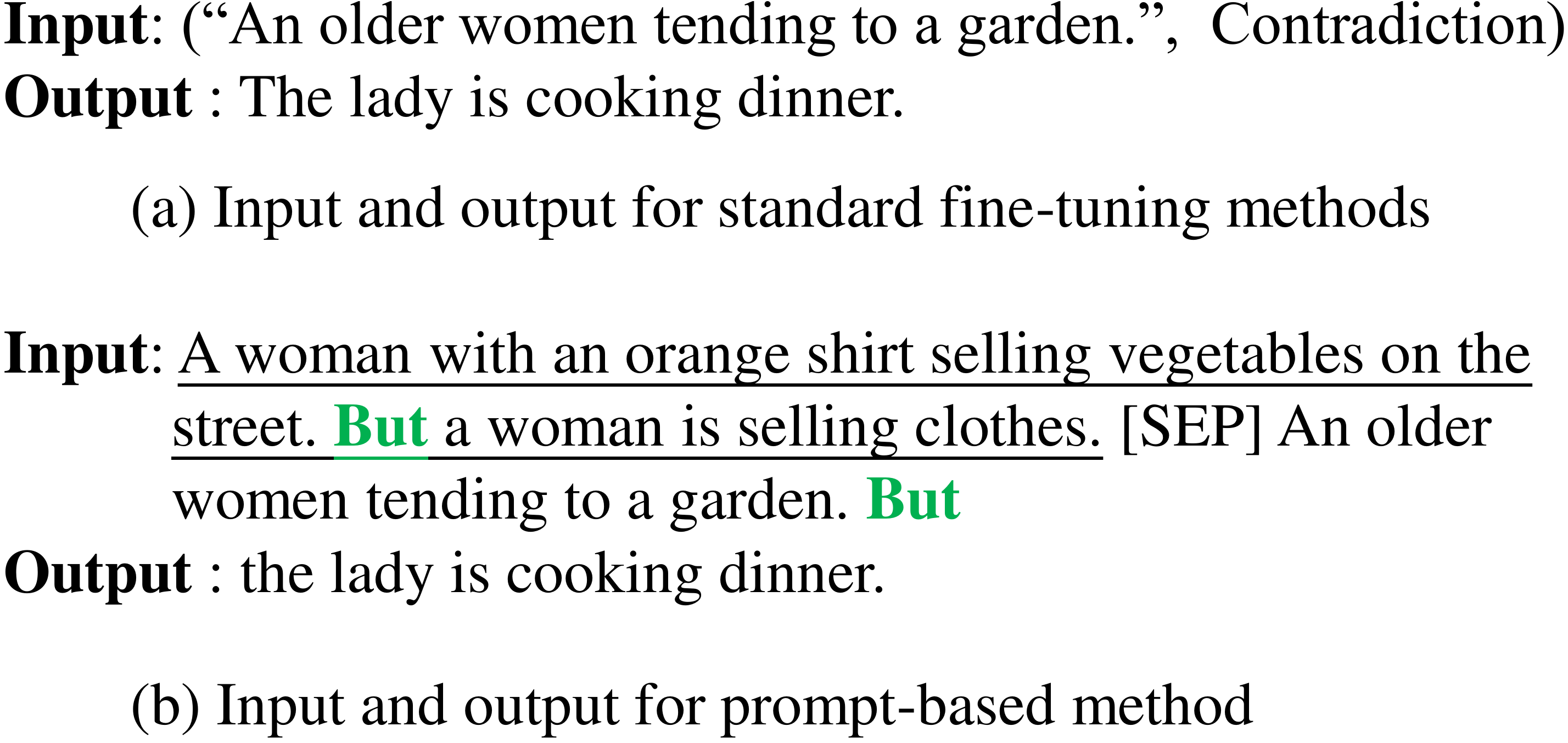}}
	\caption{The difference between how standard fine-tuning methods and prompt-based methods treat the NLI generation task. The underlined text is a demonstration. The Green words are condition-specific templates.} \label{fig:intro_prompt}
\end{figure}

However, the success of these works heavily depends on the 
SNLI dataset~\citep{DBLP:conf/emnlp/BowmanAPM15}, which is a manually created 
dataset. Thus it's a challenge to use the proposed approaches in specific domains or real-world applications which are different from the SNLI domain, 
and where there is not enough annotated data. In light of this, 
we want to explore how to improve the performance of NLI generation task 
in a few-shot setting.

Recently prompt-based learning has led to large improvements in NLP tasks under zero-shot or few-shot settings~\citep{liu2021pre,DBLP:conf/emnlp/PetroniRRLBWM19,DBLP:conf/eacl/SchickS21,DBLP:conf/emnlp/ShinRLWS20}. The main idea is to reformulate the downstream tasks as the pre-training task of pre-trained Language models (PLMs) using a prompt template. \citet{DBLP:conf/nips/BrownMRSKDNSSAA20} show that given a few demonstrations of inputs along with the prompt, GPT-3 achieves near state-of-the-art results in some SuperGLUE tasks.
Inspired by their success, we investigate how to adapt prompts and demonstrations to NLI generation task, which is reformulated as shown in \figref{fig:intro_prompt}(b).

Finding the right prompt is the first stage. 
% \citet{DBLP:conf/acl/GaoFC20} proposed to use the generative T5 model~\citep{DBLP:journals/jmlr/RaffelSRLNMZLL20} to automatically generate a lot of candidate templates. Then measure the performance for each template on development set to choose the best one. 
The problem of generating candidate templates and choosing the best one on the development set method proposed by \citet{DBLP:conf/acl/GaoFC20} is that in NLI generation task, we need to induce different templates for different conditions which makes the result on the development set less reliable,  
since we only use a subset of the data to measure each template. 
To ameliorate this, we propose a max-margin strategy to pick the template that 
receives a high score in its corresponding condition but low scores 
in all other conditions, considering the conditions in NLI task are conflicting with each other.

The second stage is to sample demonstrations. \citet{DBLP:conf/acl/GaoFC20} and \citet{DBLP:journals/corr/abs-2101-06804} simply sample them by the static cosine similarity based on models trained on related datasets (this model is regarded as a \textit{retriever}). This approach is suboptimal as: 
(1) it doesn't make use of the training data;
(2) in cases of a low-resource language, we 
do not have sufficient labeled data to train such a retriever.
To address these issues, 
we propose a dynamic method that combines 
the probability from the generator and the probability from the retriever to help the retriever fit target tasks.

In summary, our contributions are:
(1) To the best of our knowledge, we are the first to investigate NLI generation task in a few-shot setting.
(2) We propose a max-margin template selection method and a dynamic 
demonstration selection method. Combining these two methods with PLMs, 
our LM-PDD model gains 8\% absolute improvement over the standard fine-tuning models.
(3) We test our demonstration selection method on 
13 natural language classification tasks in few-shot settings, 
the results show our method has strong generality.

%Our experiments show in a few-shot setting, (1) prompt-based methods outperform standard fine-tuning methods. (2) the template chosen by the scores in all conditions outperforms the one chosen by score in one condition. (3) training the retriever together significantly improve the performance. Combining these methods together our model gains 13\% improvement compare to standard fine-tuning methods, which provides a possible way to perform NLI generation in a few-shot scenario. Finally we test the generalization ability of our methods on 13 different natural language classification tasks. The result shows our method works for a wide range of NLP tasks.

\section{Problem Setup}

\subsection{Task Formulation}

The dataset $\mathcal{D}$ consists of premises $\mathcal{P}$, conditions $\mathcal{C}=\{entailment, neutral, contradiction\}$ and hypotheses $\mathcal{H}$. Each sample $s_i$ is a tuple $(p^i, c^i, h^i)$, where $p^i \in \mathcal{P}$, $c^i \in \mathcal{C}$ and $h^i \in \mathcal{H}$. The NLI generation task is defined as: Given premise $p$ and condition $c$, predict hypothesis $h$. This formulation is the same as ~\citet{DBLP:journals/csl/StarcM17} except we assume there are only $K$
samples for per condition in training set $\mathcal{D}_{train}$ and development set $\mathcal{D}_{dev}$. This is following the few-shot settings in ~\citet{DBLP:conf/acl/GaoFC20}.

\subsection{Evaluation protocol}

Previous works show the performance of models fine-tuned on the small dataset is largely affected by the data split and hyper-parameters~\citep{DBLP:journals/corr/abs-2002-06305,DBLP:conf/iclr/0007WKWA21}. 
To address this issue, we use the same evaluation protocol in 
\citet{DBLP:conf/acl/GaoFC20}. We measure the average performance on 
5 different randomly sampled $\mathcal{D}_{train}$ and $\mathcal{D}_{dev}$ 
splits. The hyper-parameters are chosen by the performance in 
$\mathcal{D}_{dev}$ for each split.

\begin{figure*}
	\centering
	\scalebox{1.0}{\includegraphics[width=1.7\columnwidth]{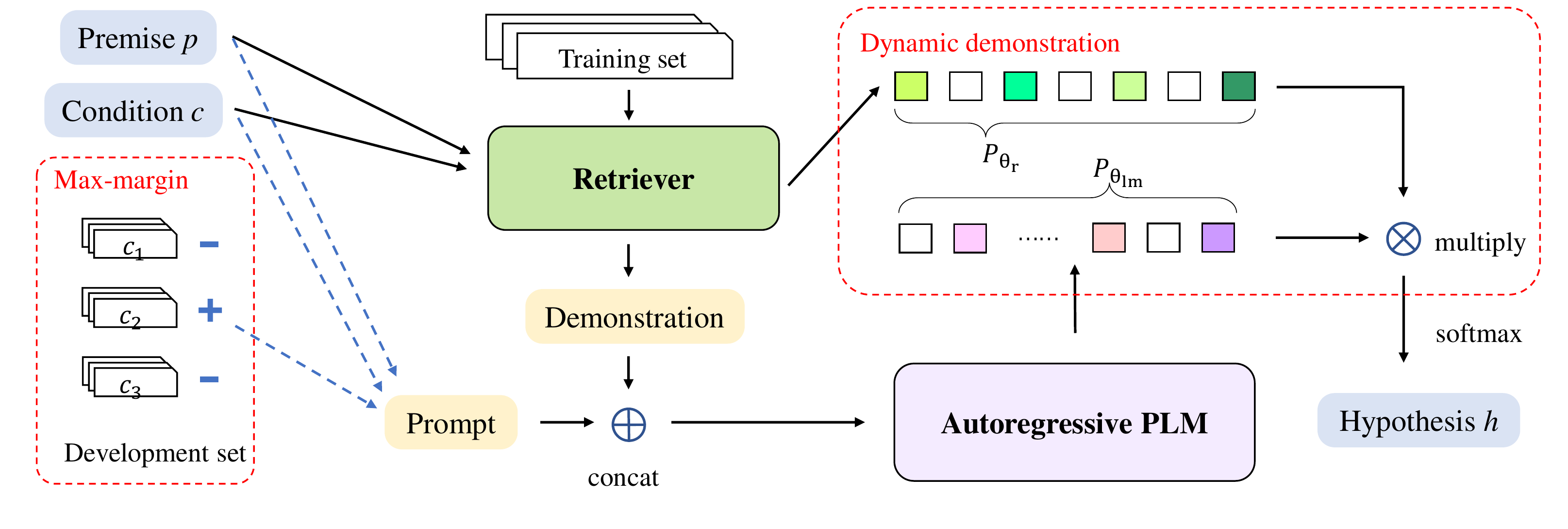}}
	\caption{An illustration of our LM-PDD. The blue dash line is the first stage: prompt template auto-generation, while the dark line stands for the dynamic demonstration training in stage two.}
	\label{fig:model}
\end{figure*}

\section{Preliminaries}

In this section, we first show how to fine-tune models with prompts and demonstrations. Then we introduce the automatic template generation method and demonstration strategy proposed in LM-BFF~\citep{DBLP:conf/acl/GaoFC20}.

\subsection{Prompt-based Fine-tuning with Demonstration}

Starting from the GPT model~\citep{radford2018improving}, pre-train and fine-tune paradigm becomes the mainstream method. To solve this task, we can use an autoregressive language model $\mathcal{L}$ (eg., GPT-2~\citep{radford2019language}, BART~\citep{DBLP:conf/acl/LewisLGGMLSZ20}). For each sample $s_i=(p^i, c^i, h^i)$, we take input $x^i$ as:
\begin{align*}
	x^i\ =\ [c^i]\ \mbox{[SEP]}\ [p^i]
\end{align*}\label{eq:fine}
$\mathcal{L}$ is fine-tuned to maximize the objective:
\begin{align}
	P(h^i | p^i, c^i) = \prod_{j}^{N} P_{\theta_{lm}} (h_{j}^i | x^i, h_{1:j-1}^i)
	\label{eq:obj}
\end{align}
However, this paradigm is hard to learn from a small amount of data. 

Prompt is a possible solution for this problem. We can reformulate the task with a template so that the input sentences are closer to the inputs during pretraining. For a sample $s_i=(p^i, contradiction, h^i)$, we can create a prompt as:
\begin{align*}
	x_{prompt}^i\ =\ [p^i]\ But\ \_\_\_
\end{align*}\label{eq:exp}
$\mathcal{L}$ will fill in the blank based on the prefix, which is the same as the pretraining task. We use the template word ``But'' here, because we hope it can direct $\mathcal{L}$ to generate the rest part with opposite meanings.
As showed in \ref{fig:intro_prompt}(b), we can also add some demonstrations in the prompt $x_{prompt}^i$ as:
\begin{align*}
	[p^{(1)}]\ But\ [h^{(1)}]\ \mbox{[SEP]}\ ...\ \mbox{[SEP]}\ [r^i]\ But\ \_\_\_
\end{align*}\label{eq:expdemo}
These demonstrations are selected from $\mathcal{D}_{train}$.

\subsection{Automatic generation of template}\label{sec:agt}

\citet{DBLP:conf/acl/GaoFC20} generate templates automatically with a T5 model~\citep{DBLP:journals/jmlr/RaffelSRLNMZLL20} for classification problems. 

For an input sample $s_i=(x_i, y_i) \in \mathcal{D}_{train}^{'}$\footnote{ $\mathcal{D}_{train}^{'}$ represents the training dataset for classification problems in their paper. We use symbol $'$ to distinguish it from the previously defined $\mathcal{D}_{train}$}, they build three different kinds of filled prompts $\mathcal{T}_g (s_i)$ whose template words are replaced by a [MASK] token.
Then a T5 model is used to fill in the [MASK] part. Because T5 is pre-trained to fill in the missing spans, there is no need to specific the number of [MASK] tokens here, which is more convenient than previous gradient-based prompt search methods~\citep{DBLP:conf/emnlp/WallaceFKGS19,DBLP:conf/emnlp/ShinRLWS20}. The goal is to find the template $\mathcal{T}$ which maximizes:
\begin{align}
	\sum_{s_i\in\mathcal{D}_{train}^{'}}log P_{T5}(\mathcal{T} | \mathcal{T}_g (s_i))
	\label{eq:decodet_old}
\end{align}

During the decoding, a wide beam search (e.g., 100) is used to obtain a large set of diverse templates. The reason why they need so many templates here is that the templates are affected by the architecture and pre-training tasks of the PLMs. Theses templates generated by T5 may not work on the $\mathcal{L}$ finally used. They fine-tuned each generated template on $\mathcal{D}_{train}^{'}$ and use $\mathcal{D}_{dev}^{'}$ to pick the best template $\mathcal{T}^{'}$:
\begin{align}
	\mathcal{T}^{'}=\mathop{\arg\max}\limits_{\mathcal{T}}\ S_{\mathcal{L}}(\mathcal{D}_{train}^{'}, \mathcal{D}_{dev}^{'}, \mathcal{T})
	\label{eq:best_old}
\end{align}
where $S_{\mathcal{L}}$ is a score function to measure performance using $\mathcal{L}$.

\subsection{Demonstration with Similarity}\label{sec:staticd}
To get the demonstrations, GPT-3 uses a random strategy to drawn from $\mathcal{D}_{train}$.  ~\citet{DBLP:conf/acl/GaoFC20} and ~\citet{DBLP:journals/corr/abs-2101-06804} use a BERT~\citep{DBLP:conf/naacl/DevlinCLT19}/RoBERTa~\citep{DBLP:journals/corr/abs-1907-11692} based retriever to sample examples with similarity. Their experiment results show (1) controlling the examples used in prompt is crucial for the model's performance; (2) using a model fine-tuned on task-related datasets is better than the original pre-trained
model.

They use SBERT~\citep{reimers-2019-sentence-bert}, which is fine-tuned on a large and diverse dataset, to build the retriever. The same SBERT model is used to encode both input samples and candidate demonstrations:
\begin{align}
	\textbf{emb}(s_i) = \mbox{SBERT}(p^i)
	\label{eq:emb}
\end{align}
The similarity is judged by cosine similarity:
\begin{align}
	Sim(s_i, s_j) = \frac{\textbf{emb}(s_i)\cdot \textbf{emb}(s_j)}{\left \| \textbf{emb}(s_i)\right \|_2\cdot \left \| \textbf{emb}(s_j)\right \|_2}
	\label{eq:sim}
\end{align}
The demonstrations examples for $s_i$ are sampled from the top 50\% instances.
%The demonstrations examples for $s_i$ are sampled from probability distribution:
%\begin{align}
%	P(s_j | s_i) = \frac{exp()}{\sum_j exp()}
%	\label{eq:simp}
%\end{align}
\section{Methodology}

The architecture of our model is depicted in Figure \ref{fig:model}. In this section, firstly we give out the hand-crafted templates. Then we introduce our prompt selection method and dynamic demonstration strategy. 

\subsection{Hand-crafted Templates}

\begin{table}[!h]
	\centering
	\small
	\begin{tabular}{l|l}
		\toprule
		\textbf{Condition} & \textbf{Template}\\
		\midrule
		Entailment  & [$p$] Then [$h$]  \\
		Neutral   & [$p$] Maybe [$h$]  \\
		Contradiction   & [$p$] But [$h$]  \\
		\bottomrule
	\end{tabular}
	\caption{Hand-crafted templates used in our experiments}
	\label{table:manua}
\end{table}
Manually defining the templates for each condition requires domain expertise knowledge. Table \ref{table:manua} shows our manually defined templates, which is different from the templates defined for NLI classification task~\citep{DBLP:conf/eacl/SchickS21}. We design them based on our knowledge, intuition, and some simple tests on the $\mathcal{L}$ we used.

\subsection{Max-margin Template Selection}

First we adapt the method in section \ref{sec:agt} so that it works on conditional generation problem in our task. For an input sample $s_i = (p^i, c^i, h^i) \in \mathcal{D}_{train}$, we build the filled prompt as:
\begin{align*}
	\mathcal{T}_g (s_i)\ =\ [p^i]\ \mbox{[MASK]}\ [h^i]
\end{align*}\label{eq:t5}
Since we need to define a different template for each condition $m \in \mathcal{C}$, equation \ref{eq:decodet_old} is modified as:
\begin{align}
	\sum_{s_i\in\mathcal{D}_{train}, c^i=m}log P_{T5}(\mathcal{T}_m | \mathcal{T}_g (s_i))
	\label{eq:decodet}
\end{align}
where $\mathcal{T}_m$ is the template used for samples whose condition is $m$. To pick the best template $\mathcal{T}_m^{'}$, we use:
\begin{align}
	\mathcal{T}_m^{'}=\mathop{\arg\max}\limits_{\mathcal{T}_m}\ S_{\mathcal{L}}(\mathcal{D}_{train}^m, \mathcal{D}_{dev}^m, \mathcal{T}_m)
	\label{eq:best}
\end{align}
where $\mathcal{D}^m=\{s_i|s_i\in\mathcal{D}, c^i=m\}$. We call this method as $top$ template selection.
 
Experiments in~\citet{DBLP:conf/acl/GaoFC20} show the size of $\mathcal{D}_{dev}$ used in equation \ref{eq:best_old} will significantly influence the quality of chosen templates, while we only use $1/|\mathcal{C}|$ samples in $\mathcal{D}_{dev}$ to measure each template in equation \ref{eq:best}.

To address this issue, we propose our max-margin template selection method. Considering the conditions in NLI generation task conflicting with each other, we can give such assumption: A good template should not get high scores in other conditions. For example, the template $[p]\ But\ [h]$ designed for ``contradiction'' are supposed to achieve a bad performance in ``entailment'' and ``neutral''. Based on this, we refine function \ref{eq:best} as:
\begin{align}
	\mathcal{T}_m^{'}=&\mathop{\arg\max}\limits_{\mathcal{T}_m}\ \sum_{k\in\mathcal{C}} d_{m,k}\cdot S_{\mathcal{L}}(\mathcal{D}_{train}^k, \mathcal{D}_{dev}^k, \mathcal{T}_m) \notag\\
	&\mbox{where,}\ d_{m,k} =
	\begin{cases} 
		1,  & \mbox{if }m=k \\
		-1, & otherwise
	\end{cases}
	\label{eq:mm}
\end{align}

\subsection{Dynamic Demonstration}

For classification problem, \citet{DBLP:conf/acl/GaoFC20} sample one example for each class. In this task, we only sample one example with the same condition as the demonstration, because templates for each condition are totally different from each other, and mixing templates as an input will mislead $\mathcal{L}$. 

We call the demonstration method in section \ref{sec:staticd} as static demonstration since the retriever is not changed in the whole experiments and the embedding for each sample is fixed. As we have mentioned in section \ref{sec:staticd}, a model fine-tuned on task-related datasets performs better, we wonder can we train the retriever based on $\mathcal{D}_{train}$ which has no similarity labels?

One solution is to annotate the similarity score for each pair. But it's unrealistic as:
(1) It's hard for a human to design the similarity boundary;
(2) the number of desired annotations is quite large even in a few-shot setting since it's square to $\left| \mathcal{D}_{train}\right|$.

Inspired by the document retrieve method from RAG~\citep{DBLP:conf/nips/LewisPPPKGKLYR020} and prompt ensemble method from~\citet{DBLP:journals/tacl/JiangXAN20}, we propose $dynamic$ demonstration method by modifying the model objective $P(h^i | p^i, c^i)$ as:
\begin{align}
	&\prod_{j}^{N} \sum_{k}P_{\theta_{r}}(s_k | s_i) P_{\theta_{lm}}(h_j^i | x_{prompt}^i, s_k, h_{1:j-1}^i)\label{eq:finalobj} \notag\\
	&\mbox{where,}\ P_{\theta_{r}}(s_k | s_i) = \frac{exp(Sim(s_i, s_k)}{\sum_{t} exp(Sim(s_i, s_t))}
\end{align}
Here $s_k, s_t\in \mathcal{D}_{train}^{c^i}$. Thus the retriever's parameters $\theta_r$ are optimized to maximize the similarity score of the input sample $s_i$ and the sample which leads to the largest generator probability of the ground truth.

During training, calculating $P_{\theta_{r}}(s_k | s_i)$ over the whole $\mathcal{D}_{train}^{c^i}$ is costly, so we do a top-k approximation. For each sample $s_i$:
(1) First we calculate $P_{\theta_{r}}(s_k | s_i)$ over $\mathcal{D}_{train}^{c^i}$ and choose the samples with top-k probability; 
(2) these chosen samples make up $\mathcal{D}_{i}$, then we calculate equation \ref{eq:finalobj}, where $s_k, s_t\in \mathcal{D}_{i}$.

During the test, we only use the retriever to sample the top-1 example as the demonstration, which is different from the decoding strategy used in RAG. Because RAG retrieves context from a corpus of 21M documents, where a lot of documents with relevant information can be found, while in a few-shot setting, the size of $\mathcal{D}_{train}$ is quite small.
\section{Experiments}
In this section we first present our experiments setup and baseline models in \secref{sec:exp} and \secref{sec:baseline} respectively. Then we show the different aspects of evaluation results and give a detailed analysis from \secref{sec:ana} to \secref{sec:ga}. According to our experiments, we want to determine these questions: First, does the maximum margin function help to automatically generate templates and does dynamic demonstration work in the few-shot setting? Second, are the sentences produced by our model reasonable? Third, how is the generalization ability of our dynamic demonstration method?

\subsection{Experiments Setup}
\noindent
\textbf{Dataset.} We evaluate our model on two English datasets: SNLI and MNLI. Because the MNLI test set doesn't have ground truth labels, we use the mismatched development set as the test set and matched development set as the development set. Different from~\citet{DBLP:conf/acl/GaoFC20} use $K=16$ for classification tasks, we set $K=200$ considering the generation task is more challenging. Detailed statistics are listed in \tabref{table:dataset}. 
\label{sec:exp}
\begin{table}[!h]
	\centering
	\small
	\begin{tabular}{l|ccc}
		\toprule
		\textbf{Dataset} & \textbf{\#Training} & \textbf{\#Development} & \textbf{\#Test}\\
		\midrule
		SNLI  & 550,152 &  10,000 & 10,000 \\
		MNLI   & 392,302 &  10,000 & 10,000 \\
		\bottomrule
	\end{tabular}
	\caption{Statistics of SNLI and MNLI dataset.}
	\label{table:dataset}
\end{table}

\begin{table*}[t!]
	\setlength\tabcolsep{4pt}
	\centering
	\small
	\begin{tabular}{l|cccccc|cccccc}
		\toprule
		\multirow{2}{*}{\textbf{Methods}} & \multicolumn{6}{c|}{\textbf{SNLI}} &\multicolumn{6}{c}{\textbf{MNLI}}  \\
		
		&\textbf{acc~(\%)}   &\textbf{B-4}  &\textbf{R-2}  &\textbf{PPL$\downarrow$}    &\textbf{Div-2}  &\textbf{Div-3}  &\textbf{acc~(\%)}   &\textbf{B-4}  &\textbf{R-2}  &\textbf{PPL$\downarrow$}     &\textbf{Div-2}  &\textbf{Div-3}   \\
		\midrule
		% \textbf{\textit{Full-shot}} & & & & & & & & & & & & & & \\
		% \midrule 
		BART-large$^\spadesuit$  & 94.17 & 8.75 & 17.63&41.91 &0.214 &0.407 &89.42 &10.55 &19.25 & 39.74& 0.365& 0.570\\
		\midrule
		% \textbf{\textit{Zero-shot}}  & & & & & & & & & & & & & & \\
		% \midrule
		Rule-base$^\clubsuit$  & 75.30 & 5.13 & 12.83 & 114.34 & 0.213& 0.402&69.07 &6.44 &16.45 &107.18 &0.289 &0.467 \\
		\midrule
		% \textbf{\textit{Few-shot}}  & & & & & & & & & & & & & & \\
		% \midrule
		AMD  & 34.75(1.91) & 0.81 & 2.19 & 678.45 & 0.097& 0.265& 34.66(3.16)&0.00 &0.25 &981.49 &0.198 & 0.547\\
		BART-large  & 66.45(8.00) & 8.63 & 15.27 &\textbf{61.93} &0.182 &0.356 &55.45(4.33) &\textbf{10.05} &16.69 &\textbf{45.35} &0.415 &0.640 \\
		\midrule
		$Prompt_{man}$ & 67.41(5.63) & 8.63 &15.58 & 67.49& 0.176& 0.343& 58.62(3.01)&9.66 &16.57 &47.46 &0.424 &0.649 \\
		$\ +static$ & 71.13(9.44) & \textbf{9.10} &\textbf{15.93} &63.95 &0.176 &0.342 &59.82(5.66) &9.90 &16.58 &47.47 &0.426 &0.650 \\
		$\ +dynamic$ & 74.01(7.89) & 8.11 &14.58 &71.36 &0.188 &0.362 &\textbf{63.33}(1.97) &9.35 &16.10 &48.87 &\textbf{0.442} &\textbf{0.674}  \\
		\midrule
		$Prompt_{top}$ & 68.04(10.01)& 8.50& 15.38& 67.79&0.176 &0.344 &59.20(9.81) &9.64 &16.45 &47.64 &0.425 &0.653 \\
		$\ +static$ & 66.79(12.23) &8.08 & 14.71& 63.57& 0.174& 0.336& 59.39(5.34)& 9.96& \textbf{16.71}& 46.39 & 0.426& 0.649\\
		$\ +dynamic$ & 73.69(4.16) & 8.10& 14.49& 68.72& 0.191& 0.367& 62.57(4.30)& 9.54& 16.29& 47.42& 0.434& 0.662\\
		\midrule
		$Prompt_{mm}$ & 69.33(6.82) & 8.51 &15.53 & 67.79& 0.176& 0.345& 59.53(4.20)&9.75 &\textbf{16.71} &47.64 &0.423 &0.647 \\
		$\ +static$ & 71.81(6.17) &8.71 & 15.49& 65.39& 0.177& 0.342& 59.78(4.33)& 9.69& 16.36& 46.60& 0.424& 0.647\\
		$\ +dynamic$ & \textbf{74.44}(4.74) & 7.97&14.42 &70.28 &\textbf{0.192} &\textbf{0.370} &62.57(2.13) &9.38 &16.04 &47.45 &0.440 &0.670 \\
		\bottomrule
	\end{tabular}
	\caption{Our main results using BART-large generator. Best results in few-shot settings are marked with \textbf{bold} font. $\spadesuit$ stands for \textbf{\textit{Full-shot}}: fine-tune with full data; $\clubsuit$  stands for \textbf{\textit{Zero-shot}}: no training data; other methods are in \textbf{\textit{Few-shot}} setting: we use $K$ = 200 (per class) for few-shot experiments. We report mean performance over 5 different split. For accuracy metric, we also report standard deviation; $man$: use manually defined prompt; $top$: auto prompt selected by maximum score; $mm$: auto prompt selected by max margin; $static$: use static SBERT as the retriever; $dynamic$: train SBERT retriever with generator; $\downarrow$:lower is better.}
	\label{table:linkprediction}
\end{table*}

\noindent
\textbf{Evaluation Metrics.} We adopt the standard text generation metrics perplexity (PPL), BLEU-4 score~\citep{papineni2002bleu} and ROUGE-2 score~\citep{lin-2004-rouge} to automatically compare generated results and reference hypotheses. Since these metrics do not consider the semantic meaning of sentences, we further use a state-of-the-art NLI classifier trained on 4 datasets~\citep{nie-etal-2020-adversarial} to calculate the accuracy of predictions, which achieves 92.6\% and 90.6\% accuracy in SNLI and MNLI mismatched datasets respectively. Good hypotheses are supposed to be not only accurate but also diverse, and therefore we use distinct n-grams normalized by the total number of generated n-grams Div-n~\citep{li-etal-2016-diversity} to measure diversity.
We also develop human evaluation in~\secref{sec:he}.

\noindent
\textbf{Implement Details.} We use BART-large as the generator. For the retriever, we deploy our experiment with the BERT-base and SBERT models. We optimize models using Adam~\citep{DBLP:journals/corr/KingmaB14}. The hyper-parameters are chosen by $\mathcal{D}_{dev}$. The max length is set to 128 for non-demonstration methods and 170 for others. We conduct all experiments on single NIVIDIA V100 16GB GPUs without half-precision. We use ROUGE-2 on the development set to choose the best models instead of the accuracy because we find the NLI classifier tends to give high accuracy even before the model is converged. More details can be found in Appendix A and B, including the generality of dynamic demonstration and generated prompts.

\subsection{Baselines}
\label{sec:baseline}
We compare our model with both unsupervised and supervised methods:

\noindent
\textbf{Rule-base:} A simple rule-based method. First, we use Stanford Stanza~\citep{qi2020stanza} to tokenize premises and get POS tags. For entailment, we simplify the premises by removing \textbf{\textit{adj}} and \textbf{\textit{adv}} words as hypothesis. For contradiction, we randomly replace one word in each premise with its antonym. If there are no candidate words, we add negative words into the sentence (e.g. ``not'' and ``don't''). For neutral, we randomly replace one word in each premise with its hyponym.

\noindent
\textbf{AMD:} A RNN-based model proposed by ~\citet{DBLP:journals/csl/StarcM17}. It learns a latent representation $\mathbf{z}$ that represents the mapping between the premise and the
label on one side, and the hypothesis on the other side.

\noindent
\textbf{BART-large:} A pre-trained language model which performances well on a range of abstractive dialogue, question answering, and summarization tasks.

\subsection{Main Results \& Analysis}
\label{sec:ana}
Our main results is shown in \tabref{table:linkprediction}. Overall, the full-shot finetuning of the BART-large model performs best except for diversity with overwhelming training consumption as a sacrifice, and the rule-base method outperforms methods in few-shot settings in terms of accuracy, while fails to perform well in other metrics. Despite the fair accuracy, the rigid rule leads the generated hypotheses less comprehensible, and the trivial edition would not provide much gain for data augmentation. In the few-shot setting, AMD is poor due to lack of pre-trained knowledge, and prompt-based methods take advantage over the most of metrics because the training and the predicting share the same LM objective. 
Generally, the B-4, R-2 and PPL are not always consistent with the accuracy result but the oscillation is slight. The increase of diversity is usually associated with the decline of B-4 and R-2, which is explainable because these metrics only consider the overlap of n-grams with one fixed reference sentence and neglect their semantic meaning. Our dynamic demonstration yields the best accuracy score in few-shot settings, achieving 8\% improvement over BART-large model in both SNLI and MNLI datasets. The diversity of our methods even exceeds BART-large trained in the full-shot scenario on MNLI, which is meaningful for the data augmentation application. 

\noindent
\textbf{Prompt Template Selection.}
When comparing different prompt generation strategies $man$, $top$ and $mm$ without demonstrations, we can find the manually designed prompt performs not as good as prompts generated by T5 model in accuracy and diversity metric, and for the accuracy of automatic generation prompts, those selected by max-margin function beat those selected by maximum score on most of metrics. This verifies our assumption that good templates of one condition get high scores in the corresponding condition while low in other conditions, and good templates narrow down the standard deviation of the accuracy.

\noindent
\textbf{Demonstration Selection.}
Under each prompt generation setting, we observe that the static demonstration improves the most of results over the original prompt except the $top$ setting, mainly due to the instability of templates generated by the maximum score. Our proposed dynamic demonstration method boosts the accuracy and diversity result with considerable gain, including the $top$ setting, which indicates that our dynamic demonstration is able to distinguish the more proper demonstration example. On the MNLI dataset, the dynamic demonstration achieves the best accuracy on manually defined templates, reflecting that in some cases the manual prompt template is simple but effective when the number of conditions is limited. In general, the dynamic demonstration shows its competitiveness no matter which prompt template selection strategy is.

\subsection{Human evaluation}
\label{sec:he}
\begin{figure}[th]
	\centering
	\scalebox{1.0}{\includegraphics[width=1\columnwidth]{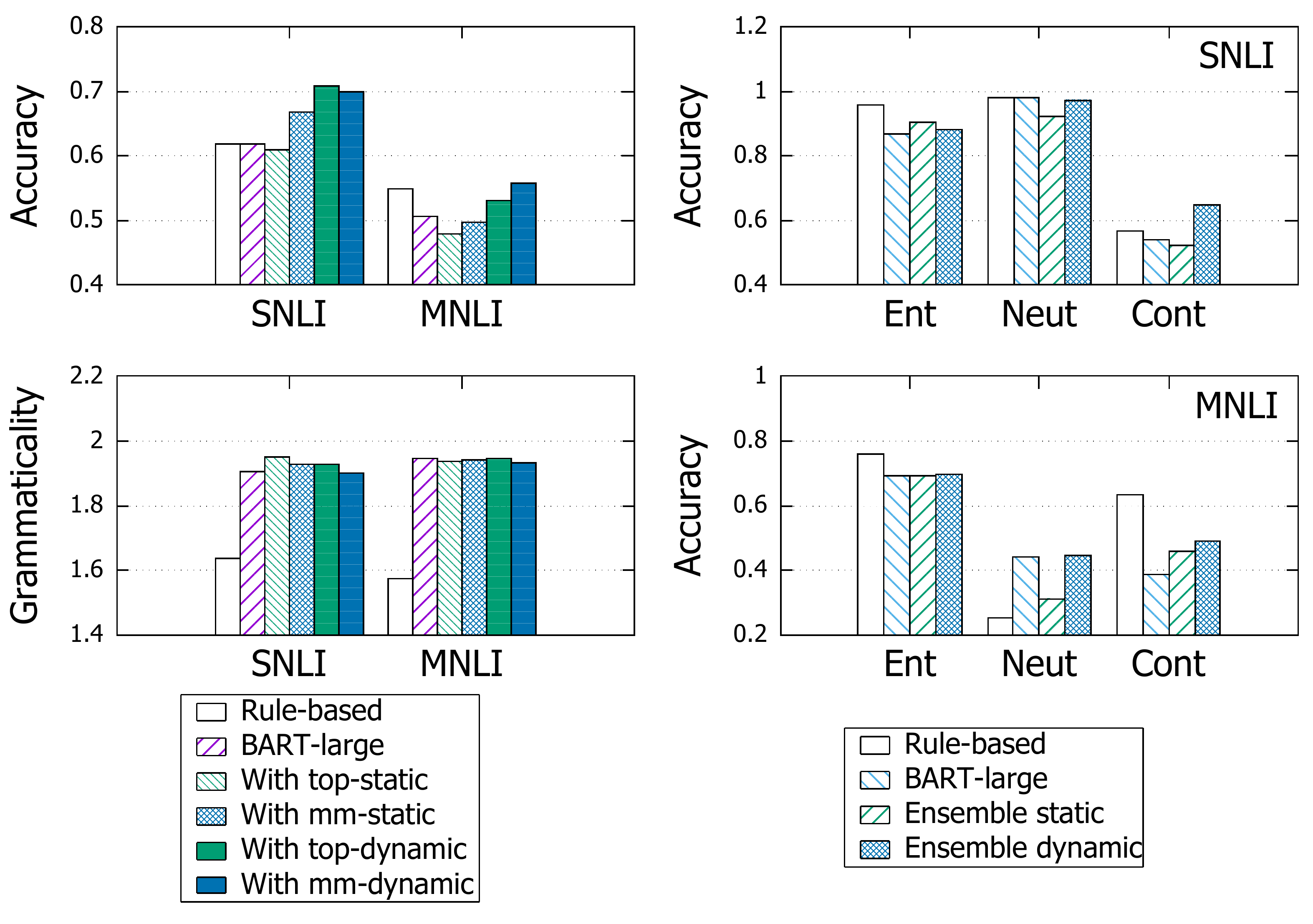}}
	\caption{First column: the human evaluation of different datasets. The same color stands for the same prompt selection. Second column: accuracy of entailment, neutral and contradiction respectively, and here we merge the results of static and dynamic demonstration of different prompt selection methods as the ensemble results.}
	\label{fig:human}
\end{figure}

In order to check the quality of the generated hypotheses, we carry out human evaluation. We recruit 5 students who are proficient in English to score the generation result of six models, with 50 samples in each condition and each dataset, and each sample is labeled by at least 3 evaluators. We examine two aspects of the quality: (1) Is the hypothesis in the right condition with the premise? (0 or 1) (2) Is the sentence reasonable grammatically? (0, 1 or 2, the higher the better). The main result is illustrated in the first two subfigures in \figref{fig:human}. For accuracy, the max-margin strategy is helpful in static demonstration, but in SNLI's dynamic demonstration things are different, maybe the power of dynamic demonstration hides the advantage of max-margin. The improvement of dynamic demonstration is always eye catching, which is consistent with auto-metric in \tabref{table:linkprediction}. For grammaticality, the rule-based method is inferior to other methods, indicating that the rule-based method owns poor readability despite the accuracy.

To further investigate the power of dynamic demonstration in different conditions, we can find that the advantage is not obvious under the scenario where these methods already obtain high accuracy, like entailment. Instead, it is more facilitative under those difficult conditions.

% 		Bart-large \\
%\textbf{Input:} A young family enjoys feeling ocean waves lap at\\ 
%their feet. \\
%\textbf{condition:} contradiction \\
%\textbf{Output:} The ocean.\\
%\textbf{Classifier output:} Entailment \\

\subsection{Case study}
The case study result is shown in \tabref{table:case}. For the static method, it finds the simple connection between ``family'' and ``child'' while fails to recognize the semantic relation between the demonstration and the premise. For the dynamic method, in addition to finding the connection between ``ocean'' and ``beach'', it also excavates the deeper pattern, which helps the PLM to generate the correct corresponding hypothesis. 
The case shows our dynamic demonstration method can find a more similar example as a demonstration to guide the generation than the static method.

\begin{table}[!h]
	\centering
	\small
	\begin{tabular}{l}
		\toprule
		SBERT + static \\
	   \textbf{Input:} \underline{A child playing with a sword at the bottom of the} \\
	   \underline{stairs. \textcolor{blue}{As} a child playing with a spoon.} [SEP] A young \\
	   family enjoys feeling ocean waves lap at their feet. \textcolor{blue}{As}\_\_\\ 
	   \textbf{condition:} Contradiction \\
	   \textbf{Output:}  The waves are crashing onto the beach..\\
	   \textbf{Classifier output:} Neutral (Bad generation)\\
		\\
		SBERT + dynamic \\
		\textbf{Input:} \underline{A middle-aged woman in a dark bathing suit and} \\\underline{her middle-aged husband in an orange hat walk cozily} \\ \underline{along the beach. \textcolor{blue}{As} there is no water.} [SEP] A young \\ family enjoys feeling ocean waves lap at their feet. \textcolor{blue}{As}\_\_\\ 
		\textbf{condition:} Contradiction \\
		\textbf{Output:}  There is no sand.\\
		\textbf{Classifier output:} Contradiction (Good generation)\\
		\bottomrule
	\end{tabular}
	\caption{Case study for the dynamic method. The underline text is demonstrations. Template words are marked as blue.}
	\label{table:case}
\end{table}
% %		Bart-large & A group of people dancing together. & A group of people is dancing. & N & E\\
%%		\midrule
%%		SBERT + dynamic & 
%\tabincell{l}{\underline{Four girls are standing together in their dance outfits.} \\\underline{\textcolor{blue}{Later,} four girls are at a dance.} [SEP] A group of \\people dancing together. \textcolor{blue}{Later,} } & A group of people are dancing. & Neutral & Entailment\\

\subsection{Generality test}
\label{sec:ga}
In order to evaluate the generality of our dynamic demonstration 
method, we implement it based on state-of-the-art few-shot approach: LM-BFF 
and develop experiments on 13 NLP classification tasks.
We use $K=10$ for all the tasks.
The LM-BFF implemented by ourselves uses the same setting in the original paper. In few-shot settings, the performance of model suffers from instability, and our implementing results are quite different from theirs\footnote{We directly run their released code from: \url{https://github.com/princeton-nlp/LM-BFF}}, so we only compare with our implementing results. As shown in table \ref{table:generazation}, our dynamic demonstration method outperforms the static method on 10 out of 13 tasks, which suggests its strong generality. 

Besides, during the test, LM-BFF samples demonstration sets from $D_{train}^{'}$ and ensemble the predictions across all sets, while for our dynamic demonstration, we abandon it because our dynamic demonstration method has the ability to tune hidden vectors on $D_{train}^{'}$ during training, and the model is expected to have the best result with the top-1 example, so there is no need to ensemble model to make the results more stable.

\begin{table}[!h]
	\centering
	\small
	\begin{tabular}{l|cc}
		\toprule
		 & \textbf{Training} & \textbf{Inference}\\
		\midrule
		LM-BFF  & 10x &  1x \\
		LM-BFF+dynamic   & 1x &  16x \\
		\bottomrule
	\end{tabular}
	\caption{Training and inference speed comparison}
	\label{table:time}
\end{table}

Table \ref{table:time} is a comparison of training and inference speed. We can see our dynamic method takes more time to train dynamic demonstration while it takes less time to do inference. The training speed for dynamic demonstration directly depends on the hyper parameter $K$, and under the few-shot setting, a small $K$ guarantees the training speed would not be too slow.

\begin{table*}[th]
	\centering
	\small
	\begin{tabular}{l|ccccccc}
		\toprule
		 & \textbf{SST-2} & \textbf{SST-5} & \textbf{MR} & \textbf{CR} & \textbf{MPQA} & \textbf{Subj} & \textbf{TREC} \\
		\midrule
		\textbf{LM-BFF} (In paper)  & 93.0(0.6) & 49.5(1.7) & 87.7(1.4) & 91.0(0.9) & 86.5(2.6) & 91.4(1.8) & 89.4(1.7) \\
		\midrule 
		\textbf{LM-BFF} (Our implement)   & \textbf{93.0}(0.5) & 50.4(1.2) & 87.3(1.9) & 90.7(1.1) & 86.2(1.8) & 90.4(2.3) & 86.2(3.9)\\
		\textbf{LM-BFF}$+dynamic$ & 91.8(0.9) & \textbf{50.6}(1.3)& \textbf{88.0}(1.1) & \textbf{91.2}(0.8) & \textbf{86.7}(1.7) & \textbf{91.2}(1.7) & \textbf{87.3}(3.0) \\
		\midrule 
		 & \textbf{CoLA} & \textbf{SNLI} & \textbf{QNLI} & \textbf{RTE} & \textbf{MRPC} & \textbf{QQP} & \textbf{-}\\
		\midrule 
		\textbf{LM-BFF} (In paper)  & 21.8(15.9) & 77.5(3.5) & 68.5(5.4) & 71.1(5.3) & 78.1(3.4) & 67.7(5.8) &\textbf{-}\\
		\midrule 
		\textbf{LM-BFF} (Our implement)   & \textbf{20.4}(14.4)& \textbf{80.5}(1.7)& 70.5(4.8) & 69.0(3.3) & 75.0(4.5) &58.6(6.4)&\textbf{-}\\
		\textbf{LM-BFF} $+ dynamic$ & 17.7(17.5) & 78.9(2.4) & \textbf{71.9}(3.7) & \textbf{72.5}(3.7) & \textbf{77.4}(3.8) & \textbf{66.4}(7.4) &\textbf{-}\\
		\bottomrule
	\end{tabular}
	\caption{Test dynamic demonstration method on 13 NLP classification tasks. }
	\label{table:generazation}
\end{table*}

\section{Related Work}

\textbf{NLI Generation}
~\citet{DBLP:journals/corr/KolesnykRR16} first proposed to serve NLI task as a generation task, while they only concentrated on entailment condition. Later ~\citet{DBLP:journals/csl/StarcM17} added two other conditions into this task to make it a conditional generation task. Our task definition follows many of their concepts except that we pay attention to few-shot settings. The previous works are all based on RNN architecture with attention mechanism, while in our work we use a stronger pre-trained language model as a baseline. There are some works exploring the inversion style of NLI generation task, which is to predict the premise based on the hypothesis and the condition. \citet{DBLP:journals/corr/KolesnykRR16} focused on entailment inversion style. They found that the model is learned to add more detailed information to the premise. \citet{DBLP:journals/corr/abs-1803-02710} used all three conditions to learn the conditional latent space over the representations of a logical antecedent of the given statement.

\noindent
\textbf{Prompt in Natural Language Generation}
Prompt-based methods is usually applied to NLG tasks by using prefix prompts together with autoregressive pre-trained LMs. \citet{radford2019language} demonstrated with prompts such as ``translate to french, [x], [z]'' or ``TL;DR'', and found that pre-trained LMs have impressive ability on generation tasks such as machine translation and summarization in the zero-shot setting. Further, \citet{DBLP:conf/nips/BrownMRSKDNSSAA20} showed that prompts also performed well in few-shot settings. \citet{DBLP:journals/corr/abs-2012-11926, DBLP:journals/corr/abs-2106-10715, DBLP:conf/acl/LiL20, DBLP:journals/corr/abs-2107-03374} explored how to adapt prompts for few-shot text summarization, machine translation, data-to-text and code generation tasks.

\noindent
\textbf{Automatic prompt search}
The automatically selected prompts consist of discrete prompts and continuous prompts (also called soft prompts). For the discrete prompts: \citet{DBLP:journals/tacl/JiangXAN20} used a text corpus to mine templates based on given training samples. Given a seed prompt, various paraphrasing methods can be exploited for generating more candidate prompts~\citep{DBLP:journals/corr/abs-2106-11520, DBLP:conf/eacl/HavivBG21}. \citet{DBLP:conf/emnlp/WallaceFKGS19} first proposed to search prompts automatically based on gradient. \citet{DBLP:conf/acl/GaoFC20} and \citet{DBLP:journals/corr/abs-2102-12206} explored how to use pre-trained T5 model to generate templates, which is more convenient compared with previous methods since it doesn't need extra corpus or modification of the model. For the continuous prompts: \citet{DBLP:conf/acl/LiL20} first proposed to use trainable continuous task-specific hidden representation vectors as prompts. There are some works making use of discrete prompts, like initializing continuous prompts with discrete prompts and hybrid prompts~\citep{DBLP:conf/naacl/ZhongFC21, DBLP:conf/naacl/QinE21, DBLP:journals/corr/abs-2103-10385}. 

\noindent
\textbf{Demonstration learning}
Demonstration learning is one of the methods to combine multi prompts. This method is first used by GPT series~\citep{radford2019language, DBLP:conf/nips/BrownMRSKDNSSAA20}. While in GPT models the demonstrations are selected randomly, researchers found that the selection with similarity would significantly improve the final performance~\citep{DBLP:conf/acl/GaoFC20, DBLP:journals/corr/abs-2101-06804}. \citet{DBLP:journals/corr/abs-2101-06804} and ~\citet{DBLP:conf/acl/KumarT21} also discovered that the order of prompts provided to the model has a great influence on the performance of the model. However, the above methods fail to select demonstrations dynamically.
\section{Conclusion}

In this paper, we investigate how to solve the natural language inference generation task in a few-shot setting. We propose LM-PDD to combine a PLM with prompts and demonstrations, including a novel template selection method, which leverages the development set for conditional generation tasks, and a dynamic demonstration method. Our methods outperform previous prompt selection and demonstration methods, achieving average 8\% accuracy over the vanilla fine-tuning method.

% Entries for the entire Anthology, followed by custom entries
\bibliography{acl2022}
\bibliographystyle{acl_natbib}
\clearpage

\renewcommand\arraystretch{1.2}
\setlength\parskip{0.1\baselineskip}
\setlength{\textfloatsep}{0.5cm}
% This is not strictly necessary, and may be commented out,
% but it will improve the layout of the manuscript,
% and will typically save some space.

%\usepackage{geometry}
%\usetikzlibrary{automata,positioning}
%\geometry{left=2.0cm, right=2.0cm, top=2.5cm, bottom=2.5cm}

\appendix

\label{appendix}

\section{Implement details}
% \label{appendix}
We take two group of hyper-parameters: learning rate=5e-5, batch size=32 and learning rate=1e-5, batch size=16. We use $\mathcal{D}_{dev}$ to chose the best one. warm-up steps are set as 10. For each model, we train it for 30 epochs in total. We evaluate it for each epoch starting from epoch 10. 

\section{Analysis of the choice of retrievers}

\begin{table}[!h]
	\centering
	\small
	\begin{tabular}{l|cccc}
		\toprule
		\multirow{2}{*} & \multicolumn{2}{c}{\textbf{SNLI(acc)}} & \multicolumn{2}{c}{\textbf{MNLI(acc)}} \\
		& \textbf{top} & \textbf{mm} & \textbf{top} & \textbf{mm}\\
		\midrule
		random &  68.49 & 69.35 & 59.02 &57.37 \\
		SBERT + static &  66.79 & 71.81 &59.39 &59.78 \\
		BERT-base + dynamic & 72.43 & 72.80 &  61.96& \textbf{62.88}\\
		SBERT + dynamic & \textbf{73.69} & \textbf{74.44} & \textbf{62.57}& 62.57\\
		\bottomrule
	\end{tabular}
	\caption{Impact of the choice of the retriever. $\textbf{random}$: randomly sample demonstrations.}
	\label{table:bert}
\end{table}

We also investigate the effect of the choice of the retriever on the performance. As showed in Table \ref{table:bert}, random retriever performs worst and dynamic SBERT retriever performs best. We also find dynamic pre-trained BERT-base outperforms static BERT, where BERT-base has similar model size with SBERT but it is not fine-tuned with extra dataset to learn a better sentence embedding. This indicates that with dynamic demonstration, the retriever can learn a better sentence embedding with training samples, so a small size of training set is enough to train a retriever well, which makes dynamic demonstration more useful in a low-resource domain.

\section{Generated prompts}

We show the prompt selected by top method and our max-margin method on SNLI and MNLI dataset in Table \ref{table:template1}, Table \ref{table:template2}, Table \ref{table:template3} and Table \ref{table:template4}

%\begin{table*}[!h]
%	\centering
%	\small
%	\begin{tabular}{l|ccccc}
%		\toprule
%		\textbf{condition} & \textbf{13} & \textbf{21} & \textbf{42} & \textbf{87} & \textbf{100}\\
%		\midrule
%		entailment  & Close up of  & A photo of & A black and white photograph of & Description: & A photo of \\
%		neutral &  Here &  Then, & At the end of the day,  & In this video,  &  And \\
%		contradiction & Now &  In the background & At the same time, & The back of & In the distance,\\
%		\bottomrule
%	\end{tabular}
%	\caption{Generated templates of top method in SNLI dataset}
%	\label{table:template1}
%\end{table*}

\begin{table*}[!h]
	\centering
	\small
	\begin{tabular}{l|ccc}
		\toprule
		\textbf{seed} & \textbf{contradiction} & \textbf{neutral} & \textbf{entailment}\\
		\midrule
		13  & Now &  Here & Close up of \\
		21   & In the background &  Then, & A photo of \\
		42   & At the same time, & At the end of the day, & A black and white photograph of \\
		87   & The back of  & In this video, & Description: \\
		100   & In the distance, &  And & A photo of \\
		\bottomrule
	\end{tabular}
	\caption{Generated templates of top method in SNLI dataset}
	\label{table:template1}
\end{table*}

%\begin{table*}[!h]
%	\centering
%	\small
%	\begin{tabular}{l|ccccc}
%		\toprule
%		\textbf{condition} & \textbf{13} & \textbf{21} & \textbf{42} & \textbf{87} & \textbf{100}\\
%		\midrule
%		entailment  & Close up of & One of & A black and white photograph of & A black and white photo of & The photo shows\\
%		neutral &  Later, &  In this photograph, & On the right  & A woman and  &  Here \\
%		contradiction & As  & At the same time & At the same time, & A close up of & Then, \\
%		\bottomrule
%	\end{tabular}
%	\caption{Generated templates of max-margin method in SNLI dataset}
%	\label{table:template2}
%\end{table*}

\begin{table*}[!h]
	\centering
	\small
	\begin{tabular}{l|ccc}
		\toprule
		\textbf{seed} & \textbf{contradiction} & \textbf{neutral} & \textbf{entailment}\\
		\midrule
		13  & As &  Later, & Close up of \\
		21   & At the same time  &  In this photograph, & One of \\
		42   & At the same time, & On the right & A black and white photograph of \\
		87   & A close up of  & A woman and & A black and white photo of \\
		100   & Then, &   Here & The photo shows \\
		\bottomrule
	\end{tabular}
	\caption{Generated templates of max-margin method in SNLI dataset}
	\label{table:template2}
\end{table*}

%\begin{table*}[!h]
%	\centering
%	\small
%	\begin{tabular}{l|ccccc}
%		\toprule
%		\textbf{condition} & \textbf{13} & \textbf{21} & \textbf{42} & \textbf{87} & \textbf{100}\\
%		\midrule
%		entailment  & The &  At the same time & That & For example, & Again,\\
%		neutral &  Although &  For the most part, & At the end of the day,  &Even  &  For the most part \\
%		contradiction & Though  & At the end of the day & Now & Although & Yet \\
%		\bottomrule
%	\end{tabular}
%	\caption{Generated templates of top method in MNLI dataset}
%	\label{table:template3}
%\end{table*}

\begin{table*}[!h]
	\centering
	\small
	\begin{tabular}{l|ccc}
		\toprule
		\textbf{seed} & \textbf{contradiction} & \textbf{neutral} & \textbf{entailment}\\
		\midrule
		13  & Though &   Although & The  \\
		21   & At the end of the day  &  For the most part, &  At the same time \\
		42   & Now & At the end of the day, & That \\
		87   & Although  & Even & For example, \\
		100   & Yet &   For the most part & Again, \\
		\bottomrule
	\end{tabular}
	\caption{Generated templates of top method in MNLI dataset}
	\label{table:template3}
\end{table*}

%\begin{table*}[!h]
%	\centering
%	\small
%	\begin{tabular}{l|ccccc}
%		\toprule
%		\textbf{condition} & \textbf{13} & \textbf{21} & \textbf{42} & \textbf{87} & \textbf{100}\\
%		\midrule
%		entailment  & In the meantime, & As you can see & At least & Finally, & ,\\
%		neutral &  So & However, & The  & Even  &  In fact  \\
%		contradiction & For some reason  & For some reason & Even though & It's not that & But in the end, \\
%		\bottomrule
%	\end{tabular}
%	\caption{Generated templates of max-margin method in MNLI dataset}
%	\label{table:template4}
%\end{table*}

\begin{table*}[!h]
	\centering
	\small
	\begin{tabular}{l|ccc}
		\toprule
		\textbf{seed} & \textbf{contradiction} & \textbf{neutral} & \textbf{entailment}\\
		\midrule
		13  & For some reason &   So & In the meantime,  \\
		21   & For some reason  & However, &  As you can see\\
		42   & Even though & The & At least \\
		87   & It's not that  & Even & Finally, \\
		100   & But in the end, &  In fact & , \\
		\bottomrule
	\end{tabular}
	\caption{Generated templates of max-margin method in MNLI dataset}
	\label{table:template4}
\end{table*}

%\begin{table*}[!h]
%	\centering
%	\small
%	\begin{tabular}{l|ccc}
%		\toprule
%		\textbf{seed} & \textbf{contradiction} & \textbf{neutral} & \textbf{entailment}\\
%		\midrule
%		13  & Now &  Here & Close up of \\
%		21   & In the background &  Then, & A photo of \\
%		42   & At the same time, & At the end of the day, & A black and white photograph of \\
%		87   & The back of  & In this video, & Description: \\
%		100   & In the distance, &  And & A photo of \\
%		\bottomrule
%	\end{tabular}
%	\caption{Statistics of SNLI and MNLI dataset.}
%	\label{table:template}
%\end{table*}
\end{document}